\documentclass{article}
\usepackage{iclr2025_conference,times}

\usepackage{amsmath,amsfonts,bm}

\def\eqref#1{equation~\ref{#1}}

\def\1{\bm{1}}

\DeclareMathAlphabet{\mathsfit}{\encodingdefault}{\sfdefault}{m}{sl}
\SetMathAlphabet{\mathsfit}{bold}{\encodingdefault}{\sfdefault}{bx}{n}

\usepackage[english]{babel}
\usepackage{amsthm,amsmath,amssymb}
\usepackage{hyperref}
\usepackage{url}
\usepackage{tabularray}

\newtheorem{theorem}{Theorem}
\newtheorem{lemma}{Lemma}

\title{Support Vector Machines Directly Minimise KL Divergence}

\author{Ivan Silajev \\Independent Researcher}

\iclrfinalcopy
\begin{document}

\maketitle

\begin{abstract}
Support vector machines (SVMs) are fundamentally grounded in geometric margin maximisation and currently lack a mainstream probabilistic interpretation. This keeps them theoretically isolated from probabilistic machine learning models that are naturally comparable and combinable. In this paper, we bridge this gap by proving that the dual objectives of both support vector classification and support vector regression are exact, constrained Kullback-Leibler (KL) divergence minimisation tasks between two Gaussian measures. We further show that this same divergence formulation recovers the exact objective function for kernel ridge regression when domain constraints are removed and $L^2$ regularisation is applied. By demonstrating that SVM duals directly minimise relative entropy, our results provide a unified information-theoretic model for standard kernel machines and lay the foundation for unifying broader kernel-based algorithms under probability theory.
\end{abstract}

\section{Introduction}\label{introduction}

Support vector machines (SVMs) are machine learning algorithms based on maximising the geometric distance of points from different classes from the separating plane in classification and minimising the distance of points from the $\varepsilon$-insensitive tube in regression \cite[Section 3]{KernelMethodsInML}. In essence, SVMs minimise the following objective function:
\begin{equation}
L(w, b) := \dfrac{1}{2} \|w\|^2 + C \sum_{i=1}^n l(y_i, \langle w, \phi(x_i) \rangle + b)
\end{equation}

where $w \in H$ and $b \in \mathbb{R}$ represent the parameters to be optimised, $x_i \in \mathcal{X}$ and $y_i \in \mathbb{R}$ represent the training data for the indices $i \in [1 , n]_{\mathbb{Z}}$ for some $n \in \mathbb{N}$. $\phi : \mathcal{X} \rightarrow H$ is a feature mapping, and the inner product $\langle . , .\rangle : H \times H \rightarrow \mathbb{R}$ defines the norm $\| .\| : H \rightarrow [0 ,\infty)$ by the identity $\| h\| := \sqrt{\langle h,h \rangle}$ for all $h \in H$. The definition of the loss function $l : \mathbb{R} \times \mathbb{R} \rightarrow \mathbb{R}$ depends on whether the objective is classification or regression. $C \in [0 ,\infty)$ is a hyper-parameter and $H$ is an inner product space. In the classification case, we have the following loss function:
\begin{equation}\label{svm-classification}
l(y, x) = l_{\text{bin}}(y, x) := \max(0, \operatorname{sign}(y)(y - x))
\end{equation}

When the target data $y_i$ take values in $\{- 1 , 1\}$, we have $\operatorname{sign}(y) = y$ and $y^2 = 1$ and the classification loss function is reduced to its familiar form $l_{\text{bin}} (y , x) = \max(0, 1 - y x)$. Meanwhile, in the regression case we have the following loss function for some $\varepsilon > 0$:
\begin{equation}\label{svm-regression}
l(y, x) = l_{\text{reg}}(y, x) := \max(0, \lvert y - x \rvert - \varepsilon)
\end{equation}

SVMs do not yet have a mainstream probabilistic and statistical interpretation. There have been attempts at formulating specific likelihood functions, the maximisation of which reproduce SVMs in \cite{ProbabilisticSVM} and ways of connecting SVMs to Gaussian processes in \cite{NIPS1999_a941493e}. Additional contributions include maximum entropy discrimination \cite{3009657.3009724}, which generalises the margin maximisation problem in classification as a constrained Kullback-Leibler divergence minimisation problem; however, it derives classical SVMs probabilistically only through the $\dfrac{1}{2} \|w\|^2$ component of the primal objective, while the hinge loss component is left as a result of a Lagrange multiplier. These attempts require theory-heavy techniques to reproduce SVMs, thus leaving the geometric interpretation to remain as the primary way of introducing the algorithm.

The lack of a general probabilistic and statistical interpretation of SVMs keeps them theoretically separate from other probabilistic machine learning algorithms. This is a problem because it deprives the field of a unified mathematical model for systems that learn from data rooted in probability theory. Machine learning algorithms that are unified under probability theory have the advantage of being comparable, combinable, and interpretable. Examples of successful work that combine machine learning algorithms under a single probabilistic theory include \cite{ICon}.

In light of these challenges, we introduce a probabilistic interpretation of SVMs by expressing their dual objectives exactly as Kullback-Leibler divergence minimisation tasks with constrained parameter spaces. This is in contrast to maximum entropy discrimination \cite{3009657.3009724}, as in our approach, the dual objective functions for both support vector classification and regression are expressed as the divergence between two Gaussian measures. We further demonstrate that this divergence serves as the exact objective function for kernel ridge regression.

We begin the paper with section \ref{preliminaries} which covers the theoretical preliminaries necessary for understanding the notation we use in our derivation, including the definition of the Kullback-Leibler divergence and its physical interpretation, as well as a subsection on defining the Hilbert space of measurable functions and a Gaussian measure on the Borel space of the said Hilbert space. Next, in section \ref{theory-section} we demonstrate the primary result and its derivation, starting with the generalised description of the primal problem for SVMs, then an end to end derivation of the generalised dual problem for SVMs and how its optimal solutions are related with the solutions of the primal problem, finally finishing off with the direct comparison of the generalised dual with the divergence between two specific Gaussian measures to show their equivalence. Furthermore, we extend our result to demonstrate that this divergence also serves as the objective function for kernel ridge regression. Before ending the section, we show how the classical dual objective for SVMs is derived exactly as a result of restricting the Hilbert space on which the Gaussian measures are defined to be finite dimensional. Finally, in section \ref{conclusion} we conclude the paper with table \ref{summary-table}, which summarises the primary differences between the machine learning algorithms analysed in this paper through the lens of our unifying theory, as well as a light overview of the type of research for which this paper will be useful. Section \ref{appendix} is reserved for showcasing the proofs of the theorems stated in this paper.

\section{Preliminaries} \label{preliminaries}

\subsection{Kullback-Leibler Divergence} \label{divergence-section}

Given a measurable space $(\Omega , \mathcal{F})$, where $\Omega$ is the state space and $\mathcal{F}$ is the event space, a sigma algebra on subsets of $\Omega$, and two probability measures $\mathbb{Q}$ and $\mathbb{P}$ defined on it such that $\mathbb{P}(A)= 0 \Rightarrow \mathbb{Q}(A) = 0$ for all $A \in \mathcal{F}$ (i.e. $\mathbb{Q}$ is absolutely continuous with respect to $\mathbb{P}$), the Kullback-Leibler divergence of $\mathbb{Q}$ with respect to $\mathbb{P}$ is given by the following formula:
\begin{equation}\label{kl-divergence}
D[\mathbb{Q}\|\mathbb{P}] := \mathbb{E}_{\mathbb{Q}}\left[ \ln \left(\frac{d\mathbb{Q}}{d\mathbb{P}}\right)\right]
\end{equation}

where $\frac{d \mathbb{Q}}{d \mathbb{P}} : \Omega \rightarrow [0 ,\infty)$ is the $\mathbb{P}$-almost surely unique Radon-Nikodym derivative of $\mathbb{Q}$ with respect to $\mathbb{P}$. When working with laws of random variables, we use the following specialised notation. Suppose $X$ and $Y$ are random variables on $(\Omega , \mathcal{F})$, then we define $\mathbb{Q}_X := \mathbb{Q} \circ X^{- 1}$ for any probability measure $\mathbb{Q}$ and $D_X [\mathbb{Q}\|\mathbb{P}] := D[\mathbb{Q}_X\|\mathbb{P}_X]$. For conditional divergences, we adopt the following notation: $D_{X | Y}[\mathbb{Q}\|\mathbb{P}](y) := D[\mathbb{Q}_{X | Y = y}\|\mathbb{P}_{X | Y = y}]$. These notational definitions can be intuitively extended for multiple variables. Finally, we mention the chain rule for divergences:
\begin{equation}
D_{X , Y}[\mathbb{Q}\|\mathbb{P}] = \mathbb{E}_{\mathbb{Q}}[D_{X | Y}[\mathbb{Q}\|\mathbb{P}](Y)] + D_Y[\mathbb{Q}\|\mathbb{P}]
\end{equation}

The divergence $D[\mathbb{Q}\|\mathbb{P}]$ has the physical interpretation of being the average number of bits of information obtained when your initial guess about the system $\mathbb{P}$ changes to $\mathbb{Q}$, given that $\mathbb{Q}$ is the actual probability distribution of the system \cite[Section 5.4]{mackay2003information}.

\subsection{Hilbert Space of Measurable Functions} \label{measurable-functions-hilbert-space}

We consider the vector space of real measurable and square integrable functions $V := \mathcal{L}^2(\mathcal{X} , \mathcal{F}_X , \mu)$ on the abstract measure space $(\mathcal{X} , \mathcal{F}_X , \mu)$. We define $\mu$-almost everywhere equality as follows for any $h , g \in V$:
\begin{equation}
h \sim g \Leftrightarrow \mu(\{x \in \mathcal{X} : h(x) \neq g(x)\}) = 0
\end{equation}

We set $H := (V / \sim) = L^2(\mathcal{X} , \mathcal{F}_X , \mu)$ as the quotient space generated by $\mu$-almost everywhere equality on $V$. We introduce an inner product $\langle . , . \rangle_H : H \times H \rightarrow \mathbb{R}$ using the following rule for any $h , g \in V$:
\begin{equation} \label{random-variable-hilbert-space-inner-product}
\langle [h], [g] \rangle_H := \int_{\mathcal{X}} h(x) g(x) d \mu(x)
\end{equation}

The space $H$ is a Hilbert space defined with the inner product $\langle . , . \rangle_H$, as its completeness is ensured by the Riesz–Fischer theorem. From here, we simplify our notation for the inner product to just the representatives of the equivalence classes by $\langle h , g \rangle_H := \langle [h], [g] \rangle_H$ for any $h , g \in V$. The Hilbert space $L^2(\mathcal{X} , \mathcal{F}_X , \mu)$ is Rockafellar-Wets decomposable, the definition of which is given in \cite[Definition 14.59]{Rockafellar1997-gr}. This fact is stated in \cite[Page 677]{Rockafellar1997-gr} and will be used in the derivation of the generalised dual problem for SVMs (i.e. for the proof of Theorem~\ref{thm:convex-conj-G-class}). Now we assume that $H$ is separable. We define the following measurable and square integrable function $Z \in L^2(\Omega \times \mathcal{X} , \mathcal{F} \otimes \mathcal{F}_X , \mathbb{P} \otimes \mu)$: 
\begin{equation}
Z(\omega , x) := m(x) + \sum_{i = 1}^{\infty} \sqrt{\lambda_i} Z_i (\omega) \phi_i (x)
\end{equation}

where $m \in H$, the variables $\phi_i \in H$ and values $\lambda_i \in \mathbb{R}$ for $i \in \mathbb{N}$ are the eigenvectors, which form an orthonormal basis of $H$, and their respective eigenvalues of a covariance operator (self-adjoint, positive semi-definite and trace class) $\Sigma : H\rightarrow H$ and $Z_i \in L^2(\Omega, \mathcal{F}, \mathbb{P}) =: \mathcal{H}$ for $i \in \mathbb{N}$ are independent standard normal variables. By the Karhunen–Loève theorem \cite[Proposition 2.18]{Da_Prato_Zabczyk_1992}, $\omega \mapsto (x \mapsto Z (\omega , x)) : \Omega \rightarrow H$ is a $H$-valued random variable for which $\langle Z , h \rangle_H \sim N (\langle m , h \rangle_H ,\langle h , \Sigma h \rangle_H)$ is Gaussian for all $h \in H$. In this case, we say $Z$ is Gaussian and we denote it as $Z \sim N_H (m , \Sigma)$ as a shorthand. Now we set a random variable $X : \Omega \rightarrow \mathcal{X}$ and set $\mu := \mathbb{P}_X = \mathbb{P} \circ X^{- 1}$. By definition, we have the following random variable:
\begin{equation}\label{measurable-gaussian}
Z_X := (\omega , \omega') \mapsto Z(\omega , X(\omega')) \in L^2(\Omega \times \Omega , \mathcal{F} \otimes \mathcal{F} , \mathbb{P} \otimes \mathbb{P})
\end{equation}

By \cite[Proposition 1.18]{Da_Prato2006-cy}, the random variable $\omega \mapsto (\omega' \mapsto Z_X (\omega ,\omega'))$ is a $\mathcal{H}$-valued Gaussian random variable such that $Z_X \sim N_{\mathcal{H}}(J m , J \Sigma J^{*})$, where $J : H \rightarrow \mathcal{H}$ is a linear operator defined by $J h := h \circ X$ and its adjoint $J^{*} : \mathcal{H} \rightarrow H$ is defined by $J^{*} h := x \mapsto \mathbb{E}[h | X = x]$. The following theorem will be useful in defining the generalised primal problem for SVMs, as it provides a rigorous way to define the Gaussian variable $Z_X$ on the diagonal set $\{(\omega , \omega') \in \Omega \times \Omega : \omega = \omega'\}$ in cases where the event has a measure of zero.

\begin{theorem}[Single Variable Evaluation Limit]\label{thm:single-variable-gaussian}
Suppose that $X$ is independent of $Z$ under $\mathbb{P}$, then define the following parameterised random variable for $n \in \mathbb{N}_0$:
\begin{equation}
S_n (X) := m(X) + \sum_{i = 1}^n \sqrt{\lambda_i} Z_i \phi_i (X) \in \mathcal{H}
\end{equation}

Then there is a $\mathbb{P}$-almost surely unique limit $Z(X) := \lim_{n \rightarrow \infty} (S_n (X)) \in \mathcal{H}$. 
\end{theorem} 

We have $\mathbb{E}[Z(X)Y | Z] = \langle Z_X , Y \rangle_{\Phi}$ $\mathbb{P}$-almost surely for when $\Phi := L^2 (\Omega , \sigma(X, Y), \mathbb{P}) \subseteq \mathcal{H}$, where the random variables $X$ and $Y$ are independent of $Z$.

\section{Theory} \label{theory-section}

\subsection{Primal Problem}

We begin by defining the primal problem for support vector machines, both classification and regression. The objective function $L : \Theta \rightarrow \mathbb{R}$ that we wish to minimise is the following:
\begin{equation}\label{primal-problem}
L(\theta) := \dfrac{1}{2} \mathbb{E}[W^2] + C \mathbb{E}[l(Y , \mathbb{E}[W Z(X)| X] + b)]
\end{equation}

where $X : \Omega \rightarrow \mathcal{X}$ is the predictor random variable, the target variable $Y : \Omega \rightarrow \mathbb{R}$ is such that $Y \in L^2(\Omega , \sigma(X, Y), \mathbb{P}) := \Phi \subseteq \mathcal{H}$, the evaluated Gaussian random variable $Z(X) \in \mathcal{H}$ is defined as in Theorem~\ref{thm:single-variable-gaussian} and the accompanying Gaussian variable $Z_X \sim N_{\Phi} (0 , K)$ for some covariance operator $K : \Phi \rightarrow \Phi$ is defined via \ref{measurable-gaussian}, where $X$ and $Y$ are independent of $Z$. The parameter is defined as $\theta := \left(W , b\right) \in L^2\left(\Omega , \sigma(Z), \mathbb{P}\right) \times \mathbb{R} =: \Theta$. The value $C \in \left[0 , \infty\right)$ is a hyper-parameter and $l : \mathbb{R} \times \mathbb{R} \rightarrow \mathbb{R}$ is a measurable function for which $x \mapsto l\left(y , x\right) : \mathbb{R} \rightarrow \mathbb{R}$ is convex for all $y \in \mathbb{R}$. For classification we set $l = l_{\text{bin}}$ as defined in \ref{svm-classification}, while for regression we set $l = l_{\text{reg}}$ as defined in \ref{svm-regression}.

Finally, the problem itself is expressed as $\inf_{\theta \in \Theta}\{L(\theta)\}$. Our goal is to derive the dual problems for both classification and regression using the weak and strong Fenchel duality theorems \cite[Theorem 4.4.2 \& 4.4.3]{Borwein2005-ky}.

\subsection{Dual Problem Derivation}

To derive the dual problem from \ref{primal-problem}, we express the loss function $L$ in a form compatible with \cite[Theorem 4.4.2]{Borwein2005-ky}: 
\begin{equation}
L(\theta) = f(\theta) + (g \circ A)(\theta)
\end{equation}

where: 
\begin{itemize}
 \item $f(\theta) = \frac{1}{2} \mathbb{E}[W^2]$ such that $f : \Theta \rightarrow \mathbb{R} \cup \{\infty\}$ is convex in the Hilbert space $\Theta$
 \item $g(V) = C \mathbb{E}[l(Y , V)]$ such that $g : \Phi \rightarrow \mathbb{R} \cup \{ \infty\}$ is convex in the Hilbert space $\Phi$
 \item $A(\theta) = \mathbb{E}[W Z(X)| X] + b$ such that $A : \Theta \rightarrow \Phi$ is a bounded linear operator on $\Theta$
\end{itemize}
The Hilbert space $\Theta$ is equipped with the inner product $\langle . , . \rangle_{\Theta} : \Theta \times \Theta \rightarrow \mathbb{R}$ defined as $\langle \left(W , b\right), \left(W^{*} , b^{*}\right) \rangle_{\Theta} := \mathbb{E}\left[W W^{*}\right] + b b^{*}$. The Hilbert space $\Phi$ is equipped with the usual inner product $\langle . , . \rangle_{\Phi} : \Phi \times \Phi \rightarrow \mathbb{R}$ on Hilbert spaces of measurable functions as defined in \ref{random-variable-hilbert-space-inner-product}. To derive the dual problem according to \cite[Theorem 4.4.2]{Borwein2005-ky} we now derive the convex conjugates of the functions $f$ and $g$ as defined in \cite[Section 4.4.1]{Borwein2005-ky} and the adjoint of $A$.

\begin{theorem}[Convex Conjugate of f]\label{thm:convex-conj-F}
The convex conjugate of $f :\Theta \rightarrow \mathbb{R} \cup \{ \infty \}$ is given as follows, where $\delta(x | A) = 0$ if $x \in A$, otherwise it is $\infty$: 
\begin{equation}
f^{*}(\theta) = \frac{1}{2}\mathbb{E}[W^2] + \delta(b|\{0\})
\end{equation} 
\end{theorem}

\begin{theorem}[Adjoint of A]\label{thm:adjoint-A}
The operator $A :\Theta \rightarrow \Phi$ is bounded and its adjoint $A^{*}: \Phi \rightarrow \Theta$ is given as follows: 
\begin{equation}
A^{*}(\alpha) = (\langle Z_X , \alpha \rangle_{\Phi} , \mathbb{E}[\alpha])
\end{equation} 
\end{theorem}

\begin{theorem}[Convex Conjugate of g: Classification] \label{thm:convex-conj-G-class}
The convex conjugate of $g :\Phi \rightarrow \mathbb{R} \cup \{ \infty\}$ when $l = l_{\text{bin}}$ as defined in \ref{svm-classification} is given as follows:
\begin{equation}\label{classification-convex-dual}
g^{*}(\alpha) = \mathbb{E}[\alpha Y] + \delta(\alpha | E_{\text{bin}})
\end{equation} 

where:
\begin{equation}\label{classification-condition} 
E_{\text{bin}} := \{\alpha \in \Phi : -\operatorname{sign}(Y) \alpha = |\alpha | \leq C \ \mathbb{P}\text{-almost surely}\} \in \mathcal{B}(\Phi)
\end{equation} 
\end{theorem}

\begin{theorem}[Convex Conjugate of g: Regression]\label{thm:convex-conj-G-reg}
The convex conjugate of $g :\Phi \rightarrow \mathbb{R} \cup \{ \infty\}$ when $l = l_{\text{reg}}$ is defined as in \ref{svm-regression} is given as follows:
\begin{equation}\label{regression-convex-dual}
g^{*}(\alpha) = \mathbb{E}[\alpha Y ] + \varepsilon \mathbb{E}[|\alpha|] + \delta(\alpha | E_{\text{reg}})
\end{equation} 

where: 
\begin{equation}\label{regression-condition} 
E_{\text{reg}} := \{\alpha \in \Phi : |\alpha | \leq C \ \mathbb{P}\text{-almost surely}\} \in \mathcal{B}(\Phi)
\end{equation}
\end{theorem}

According to \cite[Theorem 4.4.2]{Borwein2005-ky}, the dual objective function is given as follows:
\begin{equation}
L^{*}(\alpha) =: - (f^{*} \circ A^{*})(\alpha) - g^{*}(-\alpha)
\end{equation}

For classification, the dual optimisation problem is given as follows:
\begin{align}\label{classification-dual}
\sup_{\alpha \in \Phi}  \{ L^{*}(\alpha)  \} 
& = \sup_{\alpha \in \Phi} \left\{-\frac{1}{2} \mathbb{E}[\langle Z_X , \alpha \rangle_{\Phi}^2] + \mathbb{E} [\alpha Y ] - \delta(-\alpha | E_{\text{bin}}) - \delta(\mathbb{E}[\alpha]|\{0\})\right\} \\
& = \sup_{\substack{\alpha \in \Phi \\ \mathbb{E}[\alpha] = 0 \\\operatorname{sign}(Y) \alpha = |\alpha |\leq C \ \mathbb{P}\text{-a.s.}}} \left\{-\frac{1}{2} \langle \alpha , K \alpha \rangle_{\Phi} + \langle \alpha , Y \rangle_{\Phi}\right\}
\end{align}

Meanwhile, for regression, the dual optimisation problem is given as follows:

\begin{align}\label{regression-dual}
\sup_{\alpha \in \Phi}\{L^{*}(\alpha)\} & = \sup_{\alpha \in \Phi}\left\{-\frac{1}{2}\mathbb{E}[\langle Z_X , \alpha \rangle_{\Phi}^2] + \mathbb{E}[\alpha Y]- \varepsilon \mathbb{E}[ |\alpha |]  -\delta(-\alpha | E_{\text{reg}}) - \delta(\mathbb{E}[\alpha]|\{0\})\right\} \\
 & = \sup_{\substack{\alpha \in \Phi \\ \mathbb{E}[\alpha] = 0 \\ |\alpha |\leq C \ \mathbb{P}\text{-a.s.}}} \left\{-\frac{1}{2} \langle \alpha , K \alpha \rangle_{\Phi} + \langle \alpha , Y \rangle_{\Phi} - \varepsilon \mathbb{E}[|\alpha |]\right\}
\end{align}

Finally, to ensure that the conditions of \cite[Theorem 4.4.3]{Borwein2005-ky} take effect and achieve strong duality, we need to show that: 
\begin{equation}
A(\operatorname{dom}(f)) \cap \operatorname{cont}(g) \neq \emptyset
\end{equation}

Where $\operatorname{dom}(f)$ is the space of all inputs for $f : \Theta \rightarrow \mathbb{R} \cup \{\infty\}$ for which $f(\theta) < \infty$. Trivially, we have $\operatorname{dom}(f) = \Theta$. Next, the image of the linear operator $A : \Theta \rightarrow \Phi$ is, in fact, the sum of the set of constant random variables in $\Phi$ and the Cameron-Martin space $\Phi^{\frac{1}{2}} := \operatorname{Im}(K^{\frac{1}{2}})$ associated with the covariance operator $K : \Phi \rightarrow \Phi$. This is because we have the following fact using the definition of $Z(X)$ from Theorem~\ref{thm:single-variable-gaussian}:
\begin{align}
A(\theta) & = \mathbb{E}[W Z(X)| X] + b \\
 & = b + \sum_{i = 1}^{\infty} \sqrt{\lambda_i} \phi_i (X) \mathbb{E}[W Z_i] \\
 & = b + K^{\frac{1}{2}} w \in \Phi^{\frac{1}{2}} + \operatorname{span}(\mathbf{1}) \subseteq \Phi
\end{align}

for some $w \in \mathcal{H}$ such that $\langle w , \phi_i (X) \rangle_{\mathcal{H}} = \mathbb{E}[W Z_i]$ for all $i \in \mathbb{N}$. The space $\operatorname{cont}(g)$ is the space of all points in $\operatorname{dom}(g)$ where $g : \Phi \rightarrow \mathbb{R} \cup \{\infty\}$ is continuous. Trivially, one can check that $g$, regardless of whether $l = l_{\text{bin}}$ or $l = l_{\text{reg}}$, is Lipschitz continuous with the Lipschitz coefficient $C$. Therefore $\operatorname{cont}(g) = \Phi$. The Cameron-Martin space $\Phi^{\frac{1}{2}}$ is not empty, since for $0 \in \Theta$, we have $A(0) = 0 \in \Phi^{\frac{1}{2}}$, therefore, strong duality holds.

\subsection{Optimal Solution}

The optimal parameters for the primal problem $\theta = (W,b) \in \Theta$ are related to the optimal parameters of the dual problem $\alpha \in \Phi$ by the following relations:
\begin{align}
    A^{*}\alpha \in \partial f (\theta) \subseteq \Theta \label{subdiff-f}\\
    -\alpha \in \partial g (A\theta) \subseteq \Phi \label{subdiff-g}
\end{align}

where $\partial$ denotes the sub-differential operator, as defined in \cite[Section 4.2]{Borwein2005-ky}. Here, we will derive the sub-differentials to express the relationships exactly. The sub-differential of $f : \Theta \rightarrow \mathbb{R} \cup \{\infty\}$ is given trivially as $\partial f (\theta) = \{(W,0)\}$. Therefore, by Theorem~\ref{thm:adjoint-A} and \ref{subdiff-f}, we have $A^{*}(\alpha) = (\langle Z_X , \alpha \rangle_{\Phi} , \mathbb{E}[\alpha]) = (W,0) \ \mathbb{P}\text{-a.s.}$, implying that at optimality we have $\langle Z_X , \alpha \rangle_{\Phi} = W \ \mathbb{P}\text{-almost surely}$. This gives us the following equality:
\begin{equation}
A \theta = \mathbb{E}[W Z(X)|X] + b =\mathbb{E}[Z(X)\langle Z_X , \alpha \rangle_{\Phi}|X] + b = K\alpha + b
\end{equation}

For the sub-differential of $g : \Phi \rightarrow \mathbb{R} \cup \{\infty\}$, we have the following theorems:

\begin{theorem}[Sub-differential of g: Classification]\label{thm:subdiff-g-class}
The sub-differential of $g :\Phi \rightarrow \mathbb{R} \cup \{ \infty\}$ when $l = l_{\text{bin}}$ as defined in \ref{svm-classification} is given as follows:
\begin{equation}\label{classification-subdifferential}
\partial g(V) = \{\alpha \in \Phi : \alpha = -\operatorname{sign}(Y)(C\mathbb{I}\{|Y|>\operatorname{sign}(Y)V\} + |\alpha|\mathbb{I}\{Y=V\}), |\alpha| \leq C\ \mathbb{P}\text{-a.s.}\} 
\end{equation} 

\end{theorem}

\begin{theorem}[Sub-differential of g: Regression]\label{thm:subdiff-g-reg}
The sub-differential of $g :\Phi \rightarrow \mathbb{R} \cup \{ \infty\}$ when $l = l_{\text{reg}}$ is defined as in \ref{svm-regression} is given as follows:
\begin{equation}\label{regression-subdifferential}
\partial g(V) = \{\alpha \in \Phi : \alpha = -\operatorname{sign}(Y-V)(C\mathbb{I}\{|Y-V|>\varepsilon\} + |\alpha|\mathbb{I}\{|Y-V|=\varepsilon\}), |\alpha| \leq C\ \mathbb{P}\text{-a.s.}\} 
\end{equation} 
    
\end{theorem}

Solving for the classification case, given \ref{subdiff-g} and Theorem~\ref{thm:subdiff-g-class}, we have the following rule, where we define $\Delta := Y - \mathbb{E}[W Z(X)|X] = Y - K\alpha$:
\begin{equation}
    \alpha = \operatorname{sign}(Y)(C\mathbb{I}\{\operatorname{sign}(Y)(\Delta-b)>0\} + |\alpha|\mathbb{I}\{\Delta=b\})
\end{equation}

Suppose we have $Y > 0$, then if $C =\alpha$, then $\Delta \ge b$, and if $0 < \alpha < C$, then $\Delta = b$, and finally if $\alpha = 0$, we have $\Delta\le b$. Now suppose $Y < 0$, then if $\alpha = -C$, then $\Delta \le b$ and if $-C < \alpha < 0$, then $\Delta = b$, and finally if $\alpha = 0$, we have $\Delta\ge b$. Finally, if $Y = 0$, then no bounds are set and $b$ can take any value in $\mathbb{R}$. Setting the following quantities:
\begin{align}
    b^-_\text{bin} &:= \operatorname{ess} \sup\{\Delta : (Y <0\ \text{and} \ \alpha <0) \ \text{or} \ (Y>0\ \text{and} \ 0 \leq\alpha < C)\}\\
    b^+_\text{bin} &:= \operatorname{ess} \inf\{\Delta : (Y > 0\ \text{and} \ \alpha > 0) \ \text{or} \ (Y < 0 \ \text{and} \ -C < \alpha \leq 0) \}
\end{align}

where $\operatorname{ess} \sup$ and $\operatorname{ess} \inf$ are the essential supremum and infimum, respectively. The set $\mathbf{B}_\text{bin}$ of optimal values for $b$ for classification is given as follows:
\begin{equation}
   b \in\mathbf{B}_\text{bin} := \begin{cases}
       \mathbb{R} \cap [ b^-_\text{bin} , b^+_\text{bin}] &\text{if} \ b^-_\text{bin} \leq b^+_\text{bin} \\
       \varnothing &\text{otherwise}
   \end{cases} 
\end{equation}

Meanwhile, solving for the regression case, given \ref{subdiff-g} and Theorem~\ref{thm:subdiff-g-reg}, we have the following rule:
\begin{equation}
    \alpha = \operatorname{sign}(\Delta-b)(C\mathbb{I}\{|\Delta - b|>\varepsilon\} + |\alpha|\mathbb{I}\{|\Delta - b|=\varepsilon\})
\end{equation}

We have that if $\alpha = C$, then $\Delta - \varepsilon \geq b$, if $0 < \alpha < C$, then $\Delta - \varepsilon = b$, if $\alpha = 0$, then $\Delta - \varepsilon \leq b \leq \Delta + \varepsilon$, if $-C < \alpha < 0$, then $b = \Delta + \varepsilon$, and finally if $\alpha = -C$ then $b \geq \Delta + \varepsilon$. Setting the following quantities:
\begin{align}
    b^-_\text{reg} := \max(\operatorname{ess} \sup \{\Delta  + \varepsilon: \alpha <0\} ,\operatorname{ess} \sup \{\Delta  - \varepsilon: 0\leq\alpha <C\})\\
    b^+_\text{reg} := \min(\operatorname{ess} \inf \{\Delta  - \varepsilon: \alpha >0\} ,\operatorname{ess} \inf \{\Delta  + \varepsilon: -C <\alpha \leq  0\})
\end{align}

The set $\mathbf{B}_\text{reg}$ of optimal values for $b$ for regression is given as follows:
\begin{equation}
   b \in\mathbf{B}_\text{reg} := \begin{cases}
       \mathbb{R} \cap [ b^-_\text{reg} , b^+_\text{reg}] &\text{if} \ b^-_\text{reg} \leq b^+_\text{reg} \\
       \varnothing &\text{otherwise}
   \end{cases} 
\end{equation}

\subsection{Divergence Connection} \label{divergence-connection}

The objective functions of the dual objectives given by \ref{classification-dual} and \ref{regression-dual} can be reproduced exactly with the divergence between two Gaussian measures. Suppose $Z_X \sim N_{\Phi} (Y , K)$ under the probability measure $\mathbb{Q}$ such that $Y \in \Phi^{\frac{1}{2}}$, where $\Phi^{\frac{1}{2}} := \operatorname{Im}(K^{\frac{1}{2}})$ is the Cameron-Martin space, while $Z_X \sim N_{\Phi} (K \alpha , K)$ under the probability measure $\mathbb{P}$. Since the two Gaussian measures $\mathbb{Q}_{Z_X}$ and $\mathbb{P}_{Z_X}$ are mutually absolutely continuous with respect to each other, according to the Feldman–Hájek theorem \cite[Theorem 2.23]{Da_Prato_Zabczyk_1992}, we use the definition of the divergence given in \ref{divergence-section} and use the formula from \cite[Theorem 12]{quang2019regularizeddivergencescovarianceoperators} to write down the following divergence:
\begin{align}\label{divergence-svm}
D_{Z_X}[\mathbb{Q}\|\mathbb{P}] & = \frac{1}{2}\| K \alpha - Y\|_K^2 \\
 & = \frac{1}{2} \langle K \alpha , K \alpha \rangle_K - \langle K \alpha , Y \rangle_K + \frac{1}{2}\| Y\|_K^2 \\
 & = \frac{1}{2} \langle \alpha , K \alpha \rangle_{\Phi} - \langle \alpha , Y \rangle_{\Phi} + \frac{1}{2}\| Y\|_K^2
\end{align}

where the inner product $\langle . , . \rangle_K : \Phi^{\frac{1}{2}} \times \Phi^{\frac{1}{2}} \rightarrow \mathbb{R}$ is defined on $\Phi^{\frac{1}{2}}$ such that for any $h , g \in \Phi^{\frac{1}{2}}$, we have $\langle h , g \rangle_K := \langle K^{-\frac{1}{2}} h , K^{-\frac{1}{2}} g \rangle_{\Phi}$, while the respective norm is defined trivially as $\| .\|_K := \sqrt{\langle . , .\rangle_K}$.

We now have that minimising $D_{Z_X}\left[\mathbb{Q}\|\mathbb{P}\right]$ given the constraints that $\mathbb{E}\left[\alpha\right] = 0$ and $\operatorname{sign}\left(Y\right) \alpha = |\alpha | \leq C \ \mathbb{P}$-almost surely for some chosen $C > 0$ exactly reproduces the support vector classification objective derived in \ref{classification-dual}. To derive such a representation for support vector regression, we define additional random variables that produce the $L^1$ loss exactly from their divergence.

Define the random variable $R : \Omega \rightarrow \mathbb{R}$ such that $R \sim N(0 ,1/\varepsilon)$ on $\mathbb{Q}$ for some $\varepsilon > 0$ and $(R |\alpha = a) \sim N(\operatorname{sign}(a)\sqrt{2 | a |} ,1/\varepsilon)$ on $\mathbb{P}$, while $\mathbb{P}_{\alpha} = \mathbb{Q}_{\alpha}$. Then we have the following divergence:
\begin{equation}
D_{R ,\alpha}[\mathbb{Q}\|\mathbb{P}] = \frac{\varepsilon}{2}\mathbb{E}\left[\left(\operatorname{sign}(\alpha)\sqrt{2 |\alpha |}\right)^2\right] = \varepsilon \mathbb{E}[|\alpha |]
\end{equation}

By making $R$ independent from $Z_X$, we now have that minimising $D_{Z_X , R , \alpha}[\mathbb{Q}\|\mathbb{P}]$ given the constraints that $\mathbb{E}[\alpha] = 0$ and $|\alpha | \leq C \ \mathbb{P}$-almost surely for some chosen $C > 0$ exactly reproduces the support vector regression objective derived in \ref{regression-dual}.

\subsection{Kernel Ridge Regression}

Given the setting of section \ref{divergence-connection}, we now set $(R |\alpha = a) \sim N(a ,1/\varepsilon)$ on $\mathbb{P}$. This gives us the following $L^2$ regularisation divergence:
\begin{equation}
D_{R ,\alpha}[\mathbb{Q}\|\mathbb{P}] = \frac{\varepsilon}{2} \mathbb{E}[\alpha^2] = \frac{\varepsilon}{2} \langle \alpha , \alpha \rangle_{\Phi}
\end{equation}

This gives us the following objective:
\begin{align} \label{krr-divergence}
D_{Z_X , R ,\alpha}[\mathbb{Q}\|\mathbb{P}] & = \frac{1}{2} \langle \alpha , K \alpha \rangle_{\Phi} - \langle \alpha , Y \rangle_{\Phi} + \frac{1}{2}\| Y\|_K^2 + \frac{\varepsilon}{2} \langle \alpha , \alpha \rangle_{\Phi} \\
 & = \frac{1}{2} \langle \alpha , (K + \varepsilon I) \alpha \rangle_{\Phi} - \langle \alpha , Y \rangle_{\Phi} + \frac{1}{2}\| Y\|_K^2
\end{align}

The global minimiser of this objective function is given by $\alpha^{*} := (K + \varepsilon I)^{- 1} Y$. The operator $(K + \varepsilon I)^{- 1} : \Phi \rightarrow \Phi$ is properly defined on the entirety of $\Phi$ as $(K + I \varepsilon) : \Phi \rightarrow \Phi$ is a bounded operator whose eigenvalues are at least $\varepsilon$, which makes it a bijection. By the bounded inverse theorem \cite[Theorem 4.12.2]{Kreyszig1978IntroductoryFA}, $K + I \varepsilon$ is invertible. The solution $\alpha^{*}$ is precisely the solution to kernel ridge regression, and thus the divergence in \ref{krr-divergence} is an exact reproduction of the objective function for kernel ridge regression.

\subsection{Finite Dimensions}

In the setting of finite dimensional Hilbert spaces of measurable functions, two results will come useful. The first is the following fact:

\begin{lemma}[Finite Dimensional $L^2$ Spaces] \label{thm:finite-dims}
For any given random variable $X : \Omega \rightarrow \mathcal{X}$ on the probability space $(\Omega, \mathcal{F}, \mathbb{P})$ and separable and complete metric space $\mathcal{X}$, the following are equivalent statements for all $n \in \mathbb{N}$:
\begin{enumerate}    
    \item $|\operatorname{supp}(\mathbb{P}_X)| = n$
    \item $\dim(L^2(\Omega, \sigma(X), \mathbb{P})) = n$
\end{enumerate}
\end{lemma}

\begin{proof}[Proof of Lemma~\ref{thm:finite-dims}]
Assume $|\operatorname{supp}(\mathbb{P}_X)| = n \in \mathbb{N}$. Since $\mathcal{X}$ is a separable and complete metric space that generates the Borel $\sigma$-algebra $\mathcal{B}(\mathcal{X})$ on which the measure $\mathbb{P}_X$ is defined, we have that $\mathbb{P}_X(\operatorname{supp}(\mathbb{P}_X)^c) = 0$ by \cite[Lemma 1.19]{Kallenberg2021-jt}, so $\mathbb{P}(X \in \{x_1, ..., x_n\}) = 1$ and $\mathbb{P}(X = x_i) > 0$ for some values $x_i \in \mathcal{X}$ for $i \in [1, n]_\mathbb{Z}$. By the Doob-Dynkin lemma \cite[Lemma 1.14]{Kallenberg2021-jt}, for any random variable $Y \in L^2(\Omega, \sigma(X), \mathbb{P})$ there exists a measurable function $f : \mathcal{X} \rightarrow \mathbb{R}$ such that $Y = f(X) = \sum_{i=1}^n f(x_i)\mathbb{I}\{X=x_i\} \ \mathbb{P}\text{-a.s.}$. The indicator functions $\mathbb{I}\{X=x_i\} \in L^2(\Omega, \sigma(X), \mathbb{P})$ for all $i \in [1, n]_\mathbb{Z}$ form an orthogonal basis of size $n$. Therefore, $\dim(L^2(\Omega, \sigma(X), \mathbb{P})) = n$.

Now assume $\dim(L^2(\Omega, \sigma(X), \mathbb{P})) = n \in \mathbb{N}$. Suppose $|\operatorname{supp}(\mathbb{P}_X)| > n$. Since any separable and complete metric space is Hausdorff, we can form $m > n$ pairwise disjoint open $\mathcal{B}(\mathcal{X})$-measurable sets $(U_i)_{i=1}^m$ around each of the $m$ values in $\operatorname{supp}(\mathbb{P}_X)$. By definition, $\mathbb{P}_X (U_i) = \mathbb{P}(X \in U_i) > 0$ for all $i \in [1,m]_\mathbb{Z}$ and we have that $\mathbb{I}\{X \in U_i\} \in L^2(\Omega, \sigma(X), \mathbb{P})$ form an orthogonal set of $m > n$ functions. This contradicts $\dim(L^2(\Omega, \sigma(X), \mathbb{P})) = n$. Therefore, we are left with the following:
\begin{equation}
\dim(L^2(\Omega, \sigma(X), \mathbb{P})) = n \Rightarrow |\operatorname{supp}(\mathbb{P}_X)| = m \leq n \Rightarrow \dim(L^2(\Omega, \sigma(X), \mathbb{P})) = m
\end{equation}

This forces $m = n$.
\end{proof}

In the setting of \ref{primal-problem}, if we make the dimension of the Hilbert space $\Phi := L^2(\Omega, \sigma(X,Y), \mathbb{P})$ finite, by Lemma~\ref{thm:finite-dims}, the random variables $X$ and $Y$ take a finite number of values $\mathbb{P}\text{-almost surely}$, similar to a training data set. We can define values $x_i \in \mathcal{X}$ and $y_i \in \mathbb{R}$ for the index $i \in [1,n]_\mathbb{Z}$ for $n := \dim(\Phi) \in \mathbb{N}$, just as defined in Section \ref{introduction} and define a random index $I : \Omega \rightarrow [1,n]_\mathbb{Z}$ such that $X = x_I$ and $Y = y_I$ and such that the indicators $\phi_i := \mathbb{I}\{I = i\}$ form an orthogonal basis in $\Phi$. We also have $\mathbb{P}(I=i) = p_i > 0$ for all $i \in [1,n]_\mathbb{Z}$. The second fact we use in this section is stated in \cite[Section 2.2]{Bardelli_Eleonora_2017_Journal_of_Geometric_Mechanics}

\begin{lemma}[Finite Dimensional Cameron-Martin Spaces]\label{thm:finite-cam-martin}
Given a non-degenerate Gaussian variable $Z : \Omega \rightarrow H$ defined on a Hilbert space $H$ according to \cite[Section 2.3.2]{Da_Prato_Zabczyk_1992} with a covariance operator $\Sigma : H \rightarrow H$, the Cameron-Martin space $H^{\frac{1}{2}} := \operatorname{Im}(\Sigma^{\frac{1}{2}})$ equals $H$ if and only if $\dim(H) < \infty$.
\end{lemma}

Lemma~\ref{thm:finite-dims} and Lemma~\ref{thm:finite-cam-martin} grant us the ability to derive classical SVMs from their divergence formulation given in \ref{divergence-connection} as follows. Given that $\Phi$ is finite, we have $Y \in \Phi^{\frac{1}{2}} = \Phi$ according to Lemma~\ref{thm:finite-cam-martin}. The operator $K : \Phi \rightarrow \Phi$ acts as the Gram matrix on variables $\alpha \in \Phi$ as follows:
\begin{equation}
K \alpha = K \left(\sum_{i=1}^n \alpha_i \mathbb{I}\{I=i\}\right) = \sum_{i,j=1}^n K_{ij}p_j\alpha_j \mathbb{I}\{I=i\} =  \mathbb{E}[K_{IJ}\alpha_J|I]
\end{equation}

where $J$ is an independent copy of $I$ and the coefficients $K_{ij} \in \mathbb{R}$ satisfy the following conditions:
\begin{enumerate}
    \item Symmetry: $K_{ij} = K_{ji}$
    \item Positive semi-definiteness: $\langle c, Kc \rangle_\Phi = \mathbb{E}[c_I K_{IJ} c_J] \ge 0$ for all $c = c_I\in \Phi$
\end{enumerate}

Finally, we convert the divergence in \ref{divergence-svm} to its familiar form, assuming that the variable $I$ is uniformly distributed, which means $p_i = 1/n$ for all $i \in [1,n]_\mathbb{Z}$.
\begin{align}
D_{Z_X}[\mathbb{Q}\|\mathbb{P}] & = \frac{1}{2} \langle \alpha , K \alpha \rangle_{\Phi} - \langle \alpha , Y \rangle_{\Phi} + \frac{1}{2}\| Y\|_K^2 \\
& = \frac{1}{2n^2} \sum_{i,j=1}^n \alpha_i K_{ij}  \alpha_j - \frac{1}{n}\sum_{i=1}^n \alpha_i  y_i + c \\
& = \frac{1}{2} \sum_{i,j=1}^n \alpha'_i K_{ij} \alpha'_j - \sum_{i=1}^n \alpha'_i y_i + c 
\end{align}

where $\alpha'_i =  \alpha_i/n$. We also have the following for the constraints:
\begin{itemize}
    \item $\mathbb{E}[\alpha] = 0 \Leftrightarrow \sum_{i=1}^n \alpha'_i = 0$
    \item $\operatorname{sign}(Y)\alpha = |\alpha| \le C \ \mathbb{P}\text{-a.s.} \Leftrightarrow \operatorname{sign}(y_i) \alpha'_i = |\alpha'_i| \le C/n = C' \ \text{for all} \ i \in [1,n]_\mathbb{Z}$
\end{itemize}

This is exactly the classical objective function for support vector classification up to an additive constant and after rescaling the parameter $\alpha$ by $n$. Analogous results can be shown for support vector and kernel ridge regression.
    
\section{Conclusion} \label{conclusion}

In this paper, we have proven that the generalised dual objectives of SVMs are exactly constrained divergence minimisation objectives, where the objective functions are precisely the divergence between two Gaussian measures, plus a $L^1$ regularisation divergence term for regression. Moreover, we showed that this divergence also reproduces the objective function for kernel ridge regression by adding the $L^2$ regularisation divergence term and removing constraints from the parameter space. To summarise the results of this paper, we give a brief overview of the main differences between support vector classification, regression, and kernel ridge regression in table \ref{summary-table}.

The research in this paper lays foundations for building a stronger unifying information theoretic theory for machine learning algorithms such as kernel PCA, kernelised Fisher LDA and more, similar to the theoretical unification proposed in the paper \cite{ICon}.

\begin{table*}[htbp]
  \centering
  \small
  \begin{tblr}{
    width = \linewidth,
    colspec = {
      X[0.85, m, l]  
      X[1.05, m, l] 
      X[0.70, m, c] 
      X[1.40, m, l]  
    },
    row{1} = {font=\bfseries, c, m},
    rowsep = 8pt,
    colsep = 8pt,
    hlines = {0.6pt, black},
    vlines = {0.6pt, black},
    hline{1,Z} = {1pt, black},
  }
    Algorithm Name & 
    Parameter Space & 
    {Channel Conditional Mean \\ $\mathbb{E}[R \mid \alpha]$} & 
    Effective Objective Function \\
    
    Support Vector Classification & 
    $\begin{aligned}
      &\alpha \in \Phi \\
      &\mathbb{E}[\alpha] = 0 \\
      &|\alpha| \le C \ \mathbb{P}\text{-a.s.} \\
      &\operatorname{sign}(Y)\alpha = |\alpha| \ \mathbb{P}\text{-a.s.}
    \end{aligned}$ & 
    $0$ & 
    $\dfrac{1}{2} \langle \alpha, K \alpha \rangle_{\Phi} - \langle \alpha, Y \rangle_{\Phi}$ \\

    Support Vector Regression & 
    $\begin{aligned}
      &\alpha \in \Phi \\
      &\mathbb{E}[\alpha] = 0 \\
      &|\alpha| \le C \ \mathbb{P}\text{-a.s.}
    \end{aligned}$ & 
    $\operatorname{sign}(\alpha)\sqrt{2 |\alpha|}$ & 
    $\dfrac{1}{2} \langle \alpha, K \alpha \rangle_{\Phi} - \langle \alpha, Y \rangle_{\Phi} + \varepsilon \mathbb{E}[|\alpha|]$ \\

    Kernel Ridge Regression & 
    $\alpha \in \Phi$ & 
    $\alpha$ & 
    $\dfrac{1}{2} \langle \alpha, (K + \varepsilon I) \alpha \rangle_{\Phi} - \langle \alpha, Y \rangle_{\Phi}$ \\
  \end{tblr}
  \caption{Unification of Support Vector Classification, Support Vector Regression, and Kernel Ridge Regression as relative entropy minimisation problems under varied channel conditional mean and parameter spaces.}
  \label{summary-table}
\end{table*}

\section{Appendix} \label{appendix}

\begin{proof}[Proof of Theorem~\ref{thm:single-variable-gaussian}]
We have to prove that the sequence $S_n (X)$ for $n \in \mathbb{N}_0$ is a martingale and that the martingale is bounded in $L^2$. Define the filtration $\mathcal{F}_t := \sigma(X, Z_1, \ldots, Z_t)$ for $t \in \mathbb{N}_0$, where $\mathcal{F}_0 := \sigma(X)$, then we have the following for $u , t \in \mathbb{N}_0$:
\begin{align}
\mathbb{E}[S_{t + u} (X) - S_t(X)|\mathcal{F}_t] & = \mathbb{E}\left[\sum_{i = t + 1}^{t + u} \sqrt{\lambda_i} Z_i \phi_i(X)\Bigg|\mathcal{F}_t\right] \\
 & = \sum_{i = t + 1}^{t + u} \sqrt{\lambda_i} \mathbb{E}[Z_i |\mathcal{F}_t] \phi_i(X) \\
 & = \sum_{i = t + 1}^{t + u} \sqrt{\lambda_i} \mathbb{E}[Z_i] \phi_i(X) = 0
\end{align}

Thus, $S_n(X)$ is a martingale for $n \in \mathbb{N}_0$. Now we show that $S_n(X)$ is $L^2$ bounded. Firstly, $\mathbb{E}[S_0(X)^2] = \langle m , m \rangle_H < \infty$. Next we derive the expected squares of the single step increments for all $n \in \mathbb{N}_0$:
\begin{align}
\mathbb{E}[(S_{n + 1} (X) - S_n (X))^2] & = \lambda_{n + 1}\mathbb{E}[Z_{n + 1}^2] \mathbb{E}[\phi_{n + 1} (X)^2] \\
 & =\lambda_{n + 1}\|\phi_{n + 1}\|_H^2 = \lambda_{n + 1}
\end{align}

Finally we have that the following quantity is bounded:
\begin{align}
\lim_{n \rightarrow \infty}\mathbb{E}[S_n (X)^2] & = \mathbb{E}[S_0 (X)^2] + \sum_{i = 0}^{\infty} \mathbb{E}[(S_{i + 1} (X) - S_i (X))^2] \\
 & = \langle m , m \rangle_H + \operatorname{Tr}(\Sigma) < \infty
\end{align}

Therefore, we have $\sup_{n \in \mathbb{N}_0} \mathbb{E}[S_n(X)^2] < \infty$ and the martingale $S_n(X)$ is bounded in $L^2$. By \cite[Theorem 12.1.1]{Williams_1991}, we have that the limit $Z(X) := \lim_{n \rightarrow \infty} (S_n(X))$ exists and is unique $\mathbb{P}$-almost surely and in $L^2$, so $Z(X) \in \mathcal{H}$. 
\end{proof}

\begin{proof}[Proof of Theorem~\ref{thm:convex-conj-F}]
\begin{align}
f^{*}(\theta) & := \sup_{\theta' \in \Theta} \{\langle \theta', \theta \rangle_{\Theta} - f(\theta')\} \\
 & = \sup_{\theta' \in \Theta} \left\{\mathbb{E}[W' W] + b' b - \frac{1}{2} \mathbb{E}[W'^2]\right\} \\
 & = \sup_{\theta' \in \Theta} \left\{\frac{1}{2} \mathbb{E}[W^2] - \frac{1}{2}\mathbb{E}[(W - W')^2] + b' b\right\} \\
 & = \frac{1}{2} \mathbb{E}[W^2] + \delta(b|\{0\})
\end{align} 
\end{proof}

\begin{proof}[Proof of Theorem~\ref{thm:adjoint-A}]
First, we prove $A$ is bounded as follows:
\begin{align}
\| A(\theta)\|_{\Phi}^2 & = \mathbb{E}[(\mathbb{E}[W Z(X)| X] + b)^2] \\
 & \leq 2(\mathbb{E}[\mathbb{E}[W Z(X)| X]^2] + b^2) \ \text{by} \ (a + b)^2 \leq 2(a^2 + b^2) \\
 & \leq 2(\mathbb{E}[W^2] \mathbb{E}[\mathbb{E}[Z(X)^2 | X]] + b^2) \ \text{by Cauchy–Schwarz inequality} \\
 & = 2(\mathbb{E}[W^2] \mathbb{E}[Z(X)^2] + b^2) \\
 & \leq 2 \max(\mathbb{E}[Z(X)^2], 1)\|\theta\|_{\Theta}^2
\end{align}

Finally, we derive the adjoint:
\begin{align}
\langle A(W , b), \alpha \rangle_{\Phi} & = \langle \mathbb{E}[W Z(X)| X] + b , \alpha \rangle_{\Phi} \\
 & = \mathbb{E}[\alpha \mathbb{E}[W Z(X)| X]] + \mathbb{E}[\alpha] b \\
  & = \mathbb{E}[\alpha W Z(X)] + \mathbb{E}[\alpha] b  \ \text{by $X$-measurability of} \ \alpha\\
 & = \mathbb{E}[W \mathbb{E}[\alpha Z(X)| Z]] + \mathbb{E}[\alpha] b \ \text{by $Z$-measurability of} \ W \\
 & = \langle (W , b), (\langle Z_X , \alpha \rangle_{\Phi} , \mathbb{E}[\alpha]) \rangle_{\Theta} \\
 & = \langle (W , b), A^{*}(\alpha) \rangle_{\Theta}
\end{align} 
\end{proof}

\begin{proof}[Proof of Theorem~\ref{thm:convex-conj-G-class}]
We have $g(V) = C \mathbb{E}[l_{\text{bin}}(Y , V)]$, so its conjugate is given as follows:
\begin{align}
g^{*}(\alpha) & := \sup_{V \in \Phi} \{\langle V , \alpha \rangle_{\Phi} - g(V)\} \\
 & = \sup_{V \in \Phi} \{\mathbb{E}[V \alpha - C l_{\text{bin}}(Y , V)]\} \\
 & = - \inf_{V \in \Phi} \{\mathbb{E}[\gamma(\omega, V)]\} 
\end{align}

Where $\gamma(\omega, x) := C l_{\text{bin}}(Y(\omega), x) - x \alpha(\omega)$ is a Carathéodory integrand, as defined in \cite[Example 14.29]{Rockafellar1997-gr}, making it a normal integrand as defined in \cite[Definition 14.27]{Rockafellar1997-gr}. Moreover, the space of functions $\Phi$ is Rockafellar-Wets decomposable with respect to $\mathbb{P}$, as discussed in section \ref{measurable-functions-hilbert-space}. Therefore, according to \cite[Theorem 14.60]{Rockafellar1997-gr} we have the following:
\begin{equation}\label{optimal-solution-equivalence}
\inf_{V \in \Phi} \{\mathbb{E}[\gamma(\omega, V)]\} = \mathbb{E}\left[\inf_{x \in \mathbb{R}} \{\gamma(\omega, x)\}\right]
\end{equation} 

For each $\omega \in \{Y > 0\} \in \sigma(X, Y)$, we have the following:
\begin{align}
\inf_{x \in \mathbb{R}}\{\gamma(\omega, x)\} & = \inf_{x \in \mathbb{R}} \begin{cases}
 - x \alpha(\omega) &\text{if}\ Y(\omega) \leq x \\
 C Y(\omega) - x (\alpha(\omega) + C) &\text{if}\ Y(\omega) > x
\end{cases} \\
 & = \begin{cases}
 - Y(\omega) \alpha(\omega) &\text{if} - C \leq \alpha(\omega) \leq 0 \\
 - \infty &\text{otherwise}
\end{cases}
\end{align}

For each $\omega \in \{Y = 0\} \in \sigma(X, Y)$, we have the following: 
\begin{align}
\inf_{x \in \mathbb{R}}\{\gamma(\omega, x)\} & = \inf_{x \in \mathbb{R}}\{- x \alpha(\omega)\} \\
 & = \begin{cases}
 0 &\text{if}\ \alpha(\omega) = 0 \\
 -\infty & \text{otherwise}
\end{cases}
\end{align}

For each $\omega \in \{Y < 0\} \in \sigma(X, Y)$, we have the following: 
\begin{align}
\inf_{x \in \mathbb{R}}\{\gamma(\omega, x)\} & = \inf_{x \in \mathbb{R}} \begin{cases}
 - x \alpha(\omega) &\text{if}\ Y(\omega) \geq x \\
 - C Y(\omega) - x (\alpha(\omega) - C) &\text{if}\ Y(\omega) < x
\end{cases} \\
 & = \begin{cases}
 - Y(\omega) \alpha(\omega) &\text{if}\ 0 \leq \alpha(\omega) \leq C \\
 -\infty & \text{otherwise}
\end{cases}
\end{align}

Combining results, we have generally that: 
\begin{equation}
\inf_{x \in \mathbb{R}}\{\gamma(\omega, x)\} = \begin{cases}
 - Y(\omega) \alpha(\omega) &\text{if}\ -\operatorname{sign}(Y(\omega))\alpha(\omega) = |\alpha(\omega)| \le C  \\
 -\infty & \text{otherwise}
\end{cases}
\end{equation}

Using this, we can trivially define the event from \ref{classification-condition} and derive the convex dual $g^{*}(\alpha)$ as given in \ref{classification-convex-dual}. 
\end{proof}

\begin{proof}[Proof of Theorem~\ref{thm:convex-conj-G-reg}]
Set $l = l_{\text{reg}}$ as defined in \ref{svm-regression}, then repeat the proof of Theorem~\ref{thm:convex-conj-G-class} up to \ref{optimal-solution-equivalence}. We have the following for each $\omega \in \Omega$:
\begin{align}
\inf_{x \in \mathbb{R}}\{\gamma(\omega, x)\} & = \inf_{x \in \mathbb{R}} \begin{cases}
 - x \alpha(\omega) &\text{if}\ Y(\omega) - \varepsilon \leq x \leq Y(\omega) + \varepsilon \\
 - x (\alpha(\omega) - C) - C(\varepsilon + Y(\omega)) &\text{if}\ Y(\omega) + \varepsilon < x \\
 - x(\alpha(\omega) + C) - C(\varepsilon - Y(\omega)) &\text{if}\ Y(\omega) -\varepsilon > x
\end{cases} \\
 & = \begin{cases}
 - Y(\omega) \alpha(\omega) - \varepsilon |\alpha(\omega)| &\text{if}\ |\alpha(\omega)| \leq C \\
 - \infty & \text{otherwise}
\end{cases}
\end{align}

Using this, we can trivially define the event from \ref{regression-condition} and derive the convex dual $g^{*}(\alpha)$ as given in \ref{regression-convex-dual}. 
\end{proof}

\begin{proof}[Proof of Theorem~\ref{thm:subdiff-g-class}]
Using \cite[Theorem 6.5]{bùi2024interchangerulesintegralfunctions}, we have the following equivalence:
\begin{equation} \label{subderiv-as-equiv}
     \partial g (V) = \{\alpha \in \Phi : \alpha(\omega) \in \partial_x(Cl_\text{bin})(Y(\omega),V(\omega)) \ \mathbb{P}\text{-a.s.}\}
\end{equation} 

Thus, we must find $\partial_x(Cl_\text{bin})(y,x)$ for $x,y \in \mathbb{R}$. For $y > 0$, we have the following:
\begin{align}
    \partial_x(Cl_\text{bin})(y,x) & = \partial\begin{cases}
        C (y-x) &\text{if}\ y \ge x\\
        0 &\text{if} \ y<x
    \end{cases}\\
    &= \begin{cases}
        \{-C\} &\text{if} \ y > x \\
        [-C, 0] &\text{if} \ y = x \\
        \{0\} &\text{if} \ y < x
    \end{cases}
\end{align}

For $y = 0$ we have that $\partial_x(Cl_\text{bin})(y,x) = \{0\}$. Finally, we have the following for $y < 0$:
\begin{align}
    \partial_x(Cl_\text{bin})(y,x) & = \partial\begin{cases}
        C (x-y) &\text{if}\ x \ge y\\
        0 &\text{if} \ x<y
    \end{cases}\\
    &= \begin{cases}
        \{0\} &\text{if} \ x < y \\
        [0,C] &\text{if} \ x = y \\
        \{C\} &\text{if} \ x > y \\
    \end{cases}
\end{align}

Therefore, $\mathbb{P}$-almost surely, we have the following:
\begin{equation}
    \alpha \in \begin{cases}
        \{-C\} &\text{if} \ Y > V \ \text{and} \ Y>0 \\
        [-C, 0] &\text{if} \ Y = V \ \text{and} \ Y>0 \\
        \{0\} &\text{if} \ (Y < V \ \text{and} \ Y>0) \ \text{or} \ (V < Y \ \text{and} \ Y<0) \ \text{or} \ Y=0 \\
        [0,C] &\text{if} \ V = Y \ \text{and} \ Y <0 \\
        \{C\} &\text{if} \ V > Y \ \text{and} \ Y <0
    \end{cases}
\end{equation}

This means $|\alpha| \leq C \ \mathbb{P}$-almost surely. Given this case form, we can express it as the following equivalent indicator form $\mathbb{P}$-almost surely:
\begin{align}
    \alpha & = -C \mathbb{I}\{Y>V, Y>0\} + \min(0,\alpha) \mathbb{I}\{Y=V, Y>0\} \\
    & + \max(0,\alpha) \mathbb{I}\{Y=V, Y<0\} + C\mathbb{I}\{V>Y, Y<0\}\\
    & = -\operatorname{sign}(Y)(C\mathbb{I}\{\operatorname{sign}(Y)(Y-V) > 0\} + |\alpha| \mathbb{I}\{Y=V\})
\end{align}
    
\end{proof}

\begin{proof}[Proof of Theorem~\ref{thm:subdiff-g-reg}]
The equivalence \ref{subderiv-as-equiv} holds for the regression case as well. We must find we must find $\partial_x(Cl_\text{reg})(y,x)$ for $x,y \in \mathbb{R}$. We have the following:
\begin{align}
    \partial_x(Cl_\text{reg})(y,x) & = \partial\begin{cases}
        C (y-x-\varepsilon) &\text{if}\ y -x\ge \varepsilon\\
        0 &\text{if} \ |y-x|<\varepsilon\\
        C(x-y-\varepsilon) &\text{if} \ x-y \ge \varepsilon
    \end{cases}\\
    &= \begin{cases}
        \{-C\} &\text{if} \ y -x> \varepsilon  \\
        [-C, 0] &\text{if} \ y -x= \varepsilon \\
        \{0\} &\text{if} \ |y-x|<\varepsilon\\
        [0,C] &\text{if} \ x -y= \varepsilon \\
        \{C\} &\text{if} \ x -y> \varepsilon
    \end{cases}
\end{align}

Therefore, $\mathbb{P}$-almost surely, we have the following:
\begin{equation}
    \alpha \in \begin{cases}
        \{-C\} &\text{if} \ Y -V> \varepsilon  \\
        [-C, 0] &\text{if} \ Y -V= \varepsilon \\
        \{0\} &\text{if} \ |Y-V|<\varepsilon\\
        [0,C] &\text{if} \ V -Y= \varepsilon \\
        \{C\} &\text{if} \ V -Y> \varepsilon
    \end{cases}
\end{equation}

This means $|\alpha| \leq C \ \mathbb{P}$-almost surely. Given this case form, we can express it as the following equivalent indicator form $\mathbb{P}$-almost surely:
\begin{align}
    \alpha & = -C \mathbb{I}\{Y-V>\varepsilon\} + \min(0,\alpha) \mathbb{I}\{Y-V=\varepsilon\} \\
    & + \max(0,\alpha) \mathbb{I}\{V-Y = \varepsilon\} + C\mathbb{I}\{V-Y>\varepsilon\}\\
    & = -\operatorname{sign}(Y-V)(C\mathbb{I}\{|Y-V| > \varepsilon\} + |\alpha| \mathbb{I}\{|Y-V| = \varepsilon\})
\end{align}

\end{proof}

\bibliography{iclr2025_conference}
\bibliographystyle{iclr2025_conference}

\end{document}